\documentclass[conference]{IEEEtran}
\IEEEoverridecommandlockouts
\usepackage{cite}
\usepackage{amsmath,amssymb,amsfonts}
\usepackage{algorithmic}
\usepackage{graphicx}
\usepackage{textcomp}
\usepackage{xcolor}
\usepackage{stfloats}
\usepackage{multirow} 
\usepackage[hyphens]{url}
\usepackage[colorlinks=true,
            linkcolor=blue,
            citecolor=blue,
            urlcolor=blue]{hyperref}
\usepackage{breakurl}
\usepackage{subcaption}  
\def\BibTeX{{\rm B\kern-.05em{\sc i\kern-.025em b}\kern-.08em
    T\kern-.1667em\lower.7ex\hbox{E}\kern-.125emX}}
\usepackage{soul}

\begin{document}


\title{Benchmarking CNN and Transformer-Based Object Detectors for UAV Solar Panel Inspection}

\author{
\IEEEauthorblockN{
Ashen Rodrigo\textsuperscript{1},
Isuru Munasinghe\textsuperscript{1},
Pubudu Sanjeewani\textsuperscript{2},
Asanka Perera\textsuperscript{3}
}
\IEEEauthorblockA{\textsuperscript{1}Faculty of Engineering, University of Moratuwa, Katubedda, Sri Lanka}
\IEEEauthorblockA{\textsuperscript{2}STEM College, School of Computing Technologies, RMIT University, Melbourne, Australia}
\IEEEauthorblockA{\textsuperscript{3}School of Engineering, University of Southern Queensland, Brisbane, Australia}
}

\maketitle

\begin{abstract}

Timely and accurate detection of defects and contaminants in solar panels is critical for maintaining the efficiency and reliability of photovoltaic (PV) systems. While recent studies have applied deep learning to PV inspection, fair benchmarking across detector architectures and unbiased handling of class imbalance remain limited. This work presents a comprehensive benchmark of convolutional and transformer-based object detectors on UAV-captured RGB imagery of solar panels. It introduces a class-targeted augmentation strategy applied exclusively to the training split to mitigate imbalance without compromising evaluation integrity. Faster R-CNN with ResNet50 and MobileNetV3 backbones, RetinaNet with ResNet50, YOLOv5, YOLOv8, and Swin Transformer backbones integrated with Faster R-CNN (Tiny, Small, and Base variants) are evaluated. Performance is assessed using mean Average Precision (mAP) across multiple IoU thresholds, precision, recall, F1 score, and inference throughput to enable accuracy-throughput tradeoff analysis relevant to UAV deployment. Experimental results show that Faster R-CNN with a ResNet50 backbone achieves the highest localization accuracy, with mAP@0.5 of 0.893 and mAP@0.5:0.95 of 0.759, whereas the MobileNetV3 variant provides the best overall reliability balance, achieving recall of 0.745, F1-score of 0.809, and accuracy of 0.679 on the test set.

The dataset and code will be released upon acceptance of the paper.

\end{abstract}


\section{Introduction}

The efficiency of PV systems is highly dependent on the cleanliness and structural integrity of the solar panels. The accumulation of dust, dirt, and bird droppings can reduce the efficiency of energy conversion by up to 80\%, while undetected physical or electrical damage can lead to long-term degradation and safety risks~\cite{patil2017review, sayyah2014energy}. Traditional manual cleaning and even most existing robotic cleaning systems cannot assess the actual contamination or defect levels on the panels~\cite{jaiganesh2022enhancing, olorunfemi2022solar, implementation2024design, munasinghe2025ijsem}. Instead, they perform uniform cleaning across the entire surface without distinguishing between clean and contaminated regions. This approach not only leads to excessive consumption of water and electricity but also introduces unnecessary mechanical stress on panels, thereby exacerbating existing microcracks or electrical faults.

\begin{figure}[t]
    \centering
    \includegraphics[width=0.6\columnwidth]{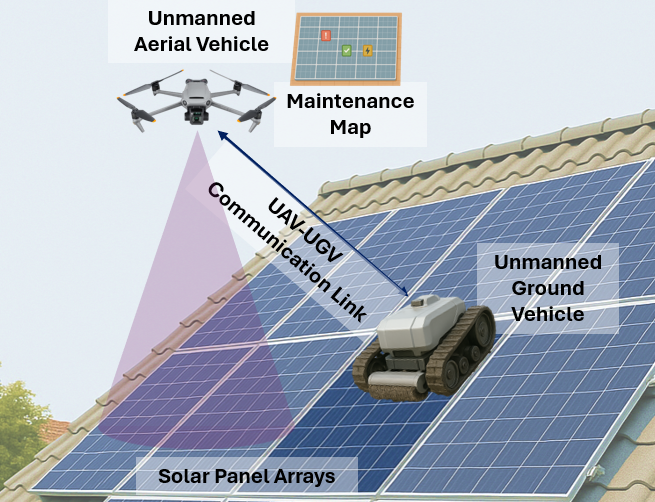}
    \caption{UAV-assisted targeted cleaning for solar panels.}
    \label{intro-image}
\end{figure}

To address these limitations, we propose an approach that integrates intelligent detection with targeted cleaning strategies. Unmanned Aerial Vehicle (UAV)-based inspection enables rapid identification of contaminated areas and isolation of physically or electrically damaged panels before cleaning, thereby improving maintenance efficiency and preventing further damage. Once contamination is detected, Unmanned Ground Vehicles (UGVs) or robotic cleaning units can selectively clean only the affected regions, avoiding panels that require repair. This proposed method reduces resource usage and operational costs while maximizing PV system output, as illustrated in Fig.~\ref{intro-image}.

Deep learning-based object detection models are pivotal for automated identification of surface contaminants, such as dust and bird droppings, as well as physical and electrical defects on solar panels~\cite{sun2023dust}. Using UAV imagery with advanced detection algorithms enables the creation of maintenance maps for UGVs, supporting the efficient and safe operation of PV systems. However, a significant challenge in this field is the limited availability of publicly accessible, well-annotated, and class-balanced solar panel defect datasets, which hinders the development and training of accurate detection models. To address these challenges and advance automated solar panel inspection, this work makes the following key contributions.

\begin{itemize}
   
    \item A comprehensive benchmark of modern object detectors spanning two-stage, single-stage, and transformer-based designs, including Faster R-CNN, RetinaNet, YOLOv5, YOLOv8, and Swin Transformer integrated Faster R-CNN, is presented.

    \item A detection performance and efficiency trade-off analysis using standardized evaluation metrics and inference throughput is provided, offering practical guidance for selecting models under real-time UAV inspection constraints.

    \item Enhancing the dataset using class-targeted augmentation to reduce class imbalance and improve minority-class representation.
\end{itemize}

The remainder of the paper is organized as follows: Section~\ref{sec2} reviews related works in solar panel defect inspection, Section~\ref{sec3} details the dataset formulation and experimental methodology, Section~\ref{sec4} presents detection benchmarks and model behavior analysis, and Section~\ref{sec5} concludes the paper with key findings. 




\section{Related Works}
\label{sec2}

Several approaches have been proposed for detecting dirt on solar panels, including sensor- and vision-based methodologies. Weight sensor systems have been used to trigger cleaning when dust accumulation causes the panel weight to exceed predefined thresholds~\cite{ronnaronglit2019cleaning}. However, these systems are susceptible to environmental vibrations and structural load variations, which can affect accuracy. Radiometric sensor-based UAV platforms have employed thermal emissivity analysis for contamination detection with reported high accuracy, although their performance may be affected by fluctuations in ambient temperature and humidity~\cite{marquez2019condition}. In~\cite{khadka2020smart}, multi-sensor prototypes were developed integrating measurements of illumination, voltage, current, temperature, humidity, and dust density to determine cleaning schedules. Nevertheless, these solutions face challenges related to sensor calibration stability and long-term operational reliability. 

Vision-based methods for solar panel inspection have utilized RGB–infrared smart camera systems with real-time image processing to enable autonomous detection and cleaning operations~\cite{yfantis2014camera, zadeh1965fuzzy}. These systems can continuously capture and analyze panel imagery, facilitating timely maintenance decisions without manual intervention. Multi-sensor fusion techniques that integrate thermal and visible-light imaging have also been applied for simultaneous hotspot identification and contaminant detection, providing a more comprehensive assessment of panel health~\cite{he2019self}. In addition, texture analysis approaches, such as those employing Gray Level Co-occurrence Matrix (GLCM) features with contrast-enhancing pre-processing, have achieved high recognition accuracy in controlled environments, enabling effective classification of dirt and soil on panel surfaces~\cite{abuqaaud2020novel}.

In~\cite{saquib2020image}, image processing was combined with dust sensors and irradiance data to estimate panel voltage using a neural network model, enabling automated cleaning triggers when output dropped below a predefined threshold. The work in~\cite{karima2023advanced} proposed a fast, self-adaptive computer vision method that uses a dust-detection camera and machine learning models to classify clean and dusty panels. In~\cite{keerthana2024image}, a comparative analysis of ANN, MLP, and DenseNet architectures for image-based dust detection showed DenseNet to achieve the highest accuracy of 98\%. The study in~\cite{qasem2016assessing} used high-resolution imagery acquired by drones and MATLAB-based algorithms to assess dust accumulation, and it was validated against electrical measurements with minimal variation. In~\cite{mamdouh2024fusion}, the image processing output was integrated with the ANN, LSTM, and SVR models to predict the performance of the solar panel under varying environmental conditions. However, the reliability of image processing thresholding methods can be adversely affected by variations in illumination, occlusions, and environmental conditions encountered in real-world scenarios.

\section{Methodology}
\label{sec3}

This section describes the dataset formulation, class definitions, augmentation strategy, and annotation preprocessing procedures used to provide unbiased performance evaluation.

\subsection{Datasets}

\subsubsection{Base Dataset and Class Definitions}

\begin{figure}[t]
    \centering
    \includegraphics[width=\columnwidth]{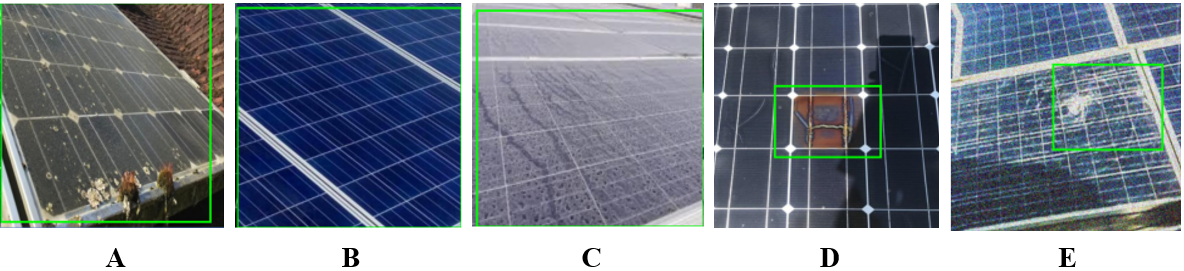}
    \caption{Representative training samples with ground truth (GT) annotations: (A) Bird droppings, (B) Clean panels, (C) Dust accumulation, (D) Electrical defects, and (E) Physical defects.}
    \label{classes}
\end{figure}

The base dataset used in this study is a publicly available solar panel defect dataset obtained from Roboflow~\cite{solarpanel_ggmtm_dataset}. It is provided with predefined training, validation, and test splits and consists of RGB images annotated with bounding boxes corresponding to multiple panel conditions and defects. The object detection task considers the five classes listed below, which represent common conditions observed in PV systems (see Fig.~\ref{classes}).

\begin{itemize}
\item \textbf{Bird Drop}: Surface contamination from bird droppings, causing partial shading and potentially hotspot formation.

\item \textbf{Clean}: Defect-free, uncontaminated panels used as the baseline for defect detection and classification.

\item \textbf{Dusty}: Accumulated dust obstructs light transmission and significantly reduces efficiency if not removed promptly.

\item \textbf{Electrical}: Visual indicators of electrical anomalies, such as localized discoloration from faults or nonuniform heating.

\item \textbf{Physical}: Visible damage on the panel, such as cracks, scratches, or deformation that can impact integrity.
\end{itemize}

\subsubsection{Dataset Augmentation Strategy}

The original training split shows an imbalanced class distribution, where Physical defects appear less frequently than other categories. This can lead to weaker detection for Physical defects compared with well-represented classes.

A targeted augmentation strategy was used to improve the balance of the training data while keeping the original evaluation splits. The strategy follows three principles.

\begin{itemize}
\item \textbf{Selective augmentation}: Only defect categories below a predefined minimum number of training images were augmented to increase their representation. Under this criterion, Physical defects were augmented to reach the minimum target.

\item \textbf{Preserving well-represented classes}: Categories that already satisfy the minimum target were left unchanged to retain the original data characteristics and avoid unnecessary augmentation.

\item \textbf{Maintaining split separation}: Augmentation was applied only to the training split. The validation and test splits remain identical to the original dataset and contain only real, unmodified images.
\end{itemize}

\begin{table}[t]
\centering
\caption{Augmentations applied to the training split using Albumentations ($p$: probability of applying the operation).}
\label{tab:augmentation_ops}
\small
\setlength{\tabcolsep}{5pt}
\renewcommand{\arraystretch}{1.15}
\begin{tabular}{lll}
\hline
\textbf{Category} & \textbf{Operation} & \textbf{Setting} \\
\hline
\multirow{3}{*}{Geometric} 
 & Horizontal flip & $p=0.5$ \\
 & Vertical flip & $p=0.5$ \\
 & Rotation & $\pm 15^\circ$, reflective padding, $p=0.5$ \\
\hline
\multirow{5}{*}{Photometric} 
 & Brightness & $\pm 20\%$, $p=0.5$ \\
 & Contrast & $\pm 20\%$, $p=0.5$ \\
 & Hue shift & $\pm 10$, $p=0.3$ \\
 & Saturation & $\pm 15\%$, $p=0.3$ \\
 & Value & $\pm 10\%$, $p=0.3$ \\
\hline
\multirow{1}{*}{Noise} 
 & Gaussian noise & variance $[10.0, 50.0]$, $p=0.2$ \\
\hline
\end{tabular}
\end{table}

The augmentation pipeline was implemented using the Albumentations library, and the applied transformations are summarized in Table~\ref{tab:augmentation_ops}. Bounding box annotations were preserved in COCO format throughout augmentation, with a minimum visibility threshold of 0.3 applied to filter boxes that become insufficiently visible after transformation. The applied operations reflect common variations in UAV inspection imagery, including viewpoint changes, camera motion, and illumination differences.









\subsubsection{Dataset Statistics, Annotation Format, and Evaluation}

Table~\ref{tab:dataset_distribution} summarizes the image distribution across classes and splits for the base dataset and the improved training split. 

\begin{table}[ht]
\caption{Image distribution per class for original and enhanced datasets.}
\centering
\begin{tabular}{lcccccc}
\hline
\multirow{2}{*}{\textbf{Class}} & \multicolumn{3}{c}{\textbf{Original Dataset}} &
\multicolumn{3}{c}{\textbf{Enhanced Dataset}} \\
\cline{2-4}\cline{5-7}
 & \textbf{Train} & \textbf{Val} & \textbf{Test} &
   \textbf{Train} & \textbf{Val} & \textbf{Test} \\
\hline
Bird Drop   & 255 & 76 & 36 & 255 & 76 & 36 \\
Clean       & 312 & 91 & 46 & 312 & 91 & 46 \\
Dusty       & 280 & 77 & 41 & 280 & 77 & 41 \\
Electrical  & 141 & 38 & 24 & 141 & 38 & 24 \\
Physical    & 116 & 30 & 15 & 140 & 30 & 15 \\
\hline
\textbf{Total} & 1104 & 312 & 162 & 1128 & 312 & 162 \\
\hline
\end{tabular}
\label{tab:dataset_distribution}
\end{table}

Annotations for the base dataset were obtained from Roboflow in COCO JSON format, containing bounding boxes $(x,y,w,h)$, class labels, and image metadata. During class-targeted augmentation, the corresponding bounding boxes were transformed together with the images using Albumentations, resulting in COCO format annotations for the augmented training samples. For YOLO-based detectors, the final training annotations (original + augmented) were converted to the YOLO format using normalized center coordinates and normalized width/height, as defined in Eq.~\ref{eq:yolo_center} and Eq.~\ref{eq:yolo_size}, where $W$ and $H$ denote image width and height, respectively. The conversion preserves a consistent class-to-index mapping across formats, and a random 5\% subset of samples was visually verified to confirm correct label mapping and bounding box placement.

\begin{equation}
x_{center} = \frac{x_{min} + \frac{w}{2}}{W}, \quad
y_{center} = \frac{y_{min} + \frac{h}{2}}{H}
\label{eq:yolo_center}
\end{equation}

\begin{equation}
w_{norm} = \frac{w}{W}, \quad
h_{norm} = \frac{h}{H}
\label{eq:yolo_size}
\end{equation}

All models were trained using the improved training split and evaluated on the unchanged validation and test splits to ensure a consistent and unbiased comparison. The enhanced dataset will be publicly released to support reproducibility and future benchmarking.

\subsection{Training Setup}
\label{sec:training}

All models were trained using PyTorch 2.0 on an NVIDIA GPU with CUDA 11.8. Most experiments used a common base configuration with the AdamW optimizer (weight decay 0.0001), an initial learning rate of 0.001, and a cosine annealing schedule. Early stopping was applied based on validation mAP@0.5 with a patience of 15 epochs. Model-specific settings, including input resolution, batch size, and the maximum number of epochs, follow the configurations summarized in Table~\ref{tab:hyperparameters_all_models} to provide stable training for each architecture.

Model performance was evaluated on the validation and test splits using standard detection metrics that capture both detection quality and deployment feasibility. mAP computed by averaging the area under the precision-recall curve across all classes is defined as
\begin{equation}
mAP = \frac{1}{N} \sum_{i=1}^{N} \int_0^1 P_i(R)\,dR.
\label{eq:map}
\end{equation}
where $P_i(R)$ denotes precision as a function of recall for class $i$, and $N$ is the number of classes. Precision and recall are computed as Eq.~\ref{eq:pr}.  $TP$ (true positives) denotes correctly detected instances, $TN$ (true negatives) denotes correctly rejected negatives, $FP$ (false positives) represents incorrect detections, and $FN$ (false negatives) corresponds to missed detections.
\begin{equation}
\text{Precision} = \frac{TP}{TP+FP}, \quad
\text{Recall} = \frac{TP}{TP+FN}
\label{eq:pr}
\end{equation}
The F1-score summarizes precision and recall using Eq.~\ref{eq:f1}.
\begin{equation}
F1 = 2 \cdot \frac{\text{Precision}\cdot\text{Recall}}{\text{Precision}+\text{Recall}}
\label{eq:f1}
\end{equation}

Deployment suitability was assessed using inference throughput in frames per second (FPS).
Localization quality was measured using Intersection over Union (IoU), defined in Eq.~\ref{eq:iou}.
\begin{equation}
IoU = \frac{|B_p \cap B_g|}{|B_p \cup B_g|}
\label{eq:iou}
\end{equation}
where $B_p$ and $B_g$ represent the predicted and GT bounding boxes, respectively. Confusion matrices and per-class metrics are also reported to analyze class-wise behavior and common misclassifications across defect categories.

Accuracy (Acc.) was computed using the standard definition as the proportion of correctly classified outcomes among all outcomes, as given in Eq.~\ref{eq:acc}. We also report the false positive rate (FPR), which measures the proportion of false positives among all actual negatives, and the false negative rate (FNR), which measures the proportion of missed positives among all actual positives to characterize error tendencies, as defined in Eq.~\ref{eq:fpr_fnr}.
\begin{equation}
Acc. = \frac{TP + TN}{TP + TN + FP + FN}.
\label{eq:acc}
\end{equation}
\begin{equation}
FPR = \frac{FP}{FP + TN}, \quad
FNR = \frac{FN}{TP + FN}.
\label{eq:fpr_fnr}
\end{equation}

The selected metrics provide a balanced view of detection performance and computational efficiency. Per-class analysis and confusion matrices further support detailed interpretation of model strengths and failure cases across different defect types.

\section{Experiments and Results}
\label{sec4}

In this section, we present a structured evaluation of object detection architectures for solar panel defect identification by comparing two-stage, single-stage, and transformer-based models, and we discuss quantitative and qualitative performance using results obtained exclusively from the test dataset.

\subsection{Training Configuration and Hyperparameter Settings}

\begin{table*}[th]
\centering
\caption{Optimal training hyperparameters used for each evaluated object detection model (all models initialized with COCO pretraining)}
\label{tab:hyperparameters_all_models}
\small
\setlength{\tabcolsep}{4pt}
\renewcommand{\arraystretch}{1.15}
\begin{tabular}{lccccccc}
\hline
\textbf{Model} & \textbf{LR} & \textbf{Scheduler} & \textbf{Optimizer} & \textbf{WD} & \textbf{Batch} & \textbf{Epochs} & \textbf{Input} \\
\hline
Faster R-CNN (ResNet50) & 0.005  & StepLR   & SGD   & 0.0005 & 2  & 30  & $640 \times 640$ \\
Faster R-CNN (MobileNetV3) & 0.005  & StepLR  & SGD & 0.0005 & 2  & 30 & $640 \times 640$ \\
RetinaNet (ResNet50) & 0.005  & StepLR  & SGD & 0.0005 & 2  & 30 & $640 \times 640$ \\
YOLOv5n & 0.01  & OneCycle & SGD   & 0.0005 & 16 & 100 & $640 \times 640$ \\
YOLOv5s & 0.01  & OneCycle & SGD   & 0.0005 & 16 & 100 & $640 \times 640$ \\
YOLOv8n & 0.01  & Cosine  & AdamW & 0.0001 & 16 & 100 & $640 \times 640$ \\
YOLOv8s & 0.01  & Cosine  & AdamW & 0.0001 & 16 & 100 & $640 \times 640$ \\
YOLOv8m & 0.01  & Cosine  & AdamW & 0.0001 & 16 & 100 & $640 \times 640$ \\
Swin-Transformer + Faster R-CNN & 0.0001 & MultiStepLR & AdamW & 0.05 & 4  & 100 & $800 \times 800$ \\
\hline
\end{tabular}
\end{table*}

We adopted consistent training configurations across all detectors to ensure fair comparison, as summarized in Table~\ref{tab:hyperparameters_all_models}. All models were initialized with COCO-pretrained weights and fine-tuned on the solar panel defect dataset. Two-stage detectors (Faster R-CNN and RetinaNet) were trained using SGD with StepLR at 0.005 learning rate for 30 epochs with 640$\times$640 inputs, while YOLO models required 100 epochs with SGD-OneCycle (YOLOv5) or AdamW-cosine (YOLOv8) optimization. The Swin Transformer configuration used AdamW with MultiStepLR, weight decay of 0.05, and 800$\times$800 inputs. Notably, Faster R-CNN models demonstrated superior convergence efficiency, achieving optimal performance in 30 epochs versus approximately 100 epochs for YOLO variants.

\subsection{Overall Quantitative Performance Evaluation}
We report the test-set performance of the evaluated detectors using complementary metrics that capture detection reliability, localization quality, and error tendencies, as summarized in Table~\ref{tab:performance_metrics_swin}. Faster R-CNN with a MobileNetV3 backbone achieved the highest F1-score of 0.809 and the lowest false negative rate of 0.255, together with competitive mAP@0.5 of 0.886 and mAP@0.5:0.95 of 0.753, indicating superior balance between sensitivity and selectivity. Faster R-CNN with a ResNet50 backbone produced the highest mAP@0.5 of 0.893 and mAP@0.5:0.95 of 0.759 with the lowest FPR of 0.107, demonstrating strong overall detection performance. RetinaNet achieved the highest IoU of 0.960, indicating strong box overlap for matched detections, although its FPR of 0.316 remained comparatively high.

\begin{table*}[th]
\centering
\caption{Comprehensive performance comparison.}
\label{tab:performance_metrics_swin}

\small
\setlength{\tabcolsep}{5pt}
\renewcommand{\arraystretch}{1.2}
\begin{tabular}{lccccccccc}
\hline
\textbf{Model} & \textbf{Precision} & \textbf{Recall} & \textbf{F1} & \textbf{Acc.} & \textbf{mAP@0.5} & \textbf{mAP@0.5:0.95} & \textbf{IoU} & \textbf{FPR} & \textbf{FNR} \\
\hline
\multicolumn{10}{l}{\textit{Two-Stage Detectors}} \\
Faster R-CNN (ResNet50) & 0.893 & 0.732 & 0.805 & 0.673 & \textbf{0.893} & \textbf{0.759} & 0.944 & \textbf{0.107} & 0.268 \\
Faster R-CNN (MobileNetV3) & 0.886 & \textbf{0.745} & \textbf{0.809} & \textbf{0.679} & 0.886 & 0.753 & 0.946 & 0.114 & \textbf{0.255} \\
RetinaNet (ResNet50) & 0.684 & 0.732 & 0.707 & 0.547 & 0.684 & 0.581 & \textbf{0.960} & 0.316 & 0.268 \\
\hline
\multicolumn{10}{l}{\textit{Single-Stage YOLO Detectors}} \\
YOLOv5 Nano (5n) & 0.887 & 0.682 & 0.771 & 0.546 & 0.809 & 0.621 & 0.854 & 0.286 & 0.301 \\
YOLOv5 Small (5s) & 0.872 & 0.712 & 0.784 & 0.543 & 0.810 & 0.627 & 0.852 & 0.308 & 0.285 \\
YOLOv8 Nano (8n) & 0.834 & 0.706 & 0.765 & 0.575 & 0.808 & 0.656 & 0.864 & 0.246 & 0.293 \\
YOLOv8 Small (8s) & \textbf{0.918} & 0.714 & 0.803 & 0.606 & 0.830 & 0.687 & 0.869 & 0.222 & 0.268 \\
YOLOv8 Medium (8m) & 0.859 & 0.693 & 0.767 & 0.542 & 0.811 & 0.654 & 0.860 & 0.270 & 0.322 \\
\hline
\multicolumn{10}{l}{\textit{Transformer-Based Models}} \\
Swin-Transformer + Faster R-CNN & 0.841 & 0.687 & 0.756 & 0.592 & 0.789 & 0.651 & 0.921 & 0.159 & 0.313 \\
\hline
\end{tabular}
\end{table*}

A comparison across model families highlights distinct operating characteristics. Two-stage detectors consistently produced the strongest mAP values, supporting their suitability when maximizing detection performance is the primary objective. Single-stage YOLO models provided competitive mAP and F1 values while maintaining high precision. In particular, YOLOv8 Small achieved the highest precision of 0.918 and a low false positive rate of 0.222, supporting deployment settings where minimizing false alarms is emphasized. The Swin Transformer-backbone Faster R-CNN configuration achieved an mAP@0.5 of 0.789 and an mAP@0.5:0.95 of 0.651, remaining below the best CNN-backbone Faster R-CNN variants under the current training configuration and dataset scale. Performance is expected to improve with larger and more diverse training data that better supports transformer-based representation learning.

\subsection{Confusion Matrix Analysis}

\begin{figure*}[t!]
    \centering
    \includegraphics[width=0.8\textwidth]{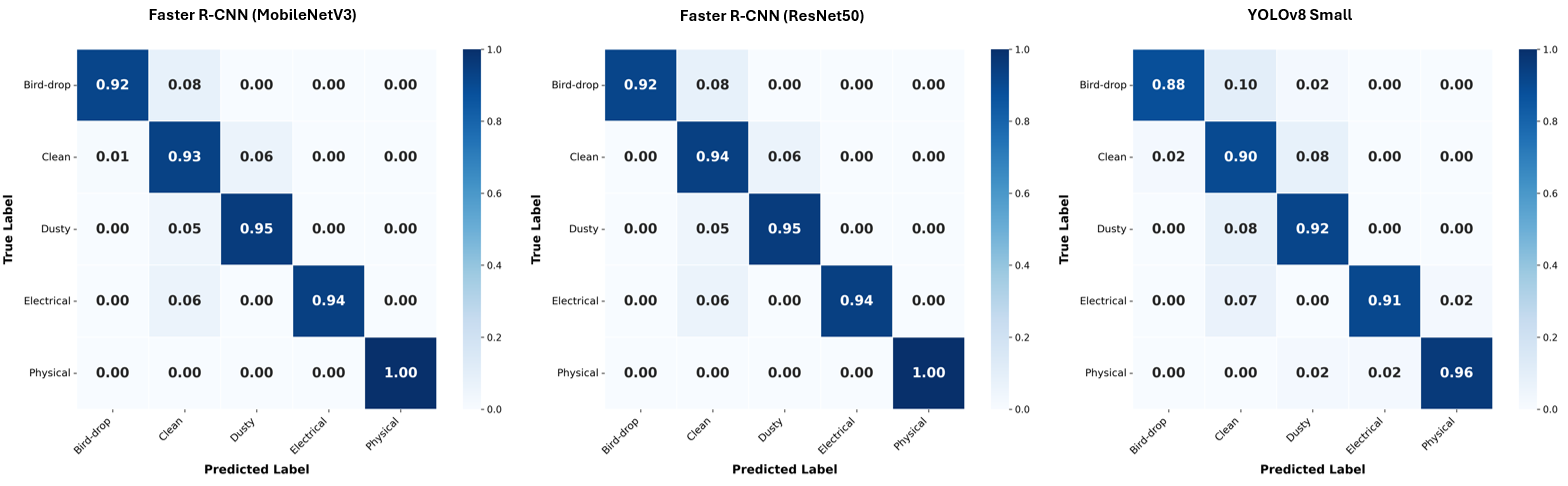}
    \caption{Confusion matrices of the best-performing three models evaluated on the test dataset.}
    \label{confusion_matrix}
\end{figure*}

The confusion matrices in Fig.~\ref{confusion_matrix} show strong diagonal dominance, indicating robust overall classification performance. The primary confusion occurs between Clean and Dusty categories (6-8\% bidirectional misclassification), attributed to visual similarities under varying lighting and gradual dust accumulation. Bird Drop exhibits 8-10\% misclassification as Clean due to subtle early-stage deposits, while 5-6\% of Dusty samples are confused with Bird Drop owing to similar visual patterns. Minor Electrical-to-Clean confusion (6-7\%) reflects the subtle appearance of early electrical defects before thermal signatures develop.
Physical Damage achieves near-perfect classification which is greater than 96\% accuracy across all models, as structural defects present distinctive visual features. Electrical Damage also performs strongly (91-94\%), benefiting from characteristic thermal signatures. These results suggest future improvements should focus on distinguishing visually similar categories, particularly Clean versus Dusty defects.

\subsection{Visual Prediction Analysis and Training Convergence}

\begin{figure*}[hb!]
    \centering
    \includegraphics[width=\textwidth]{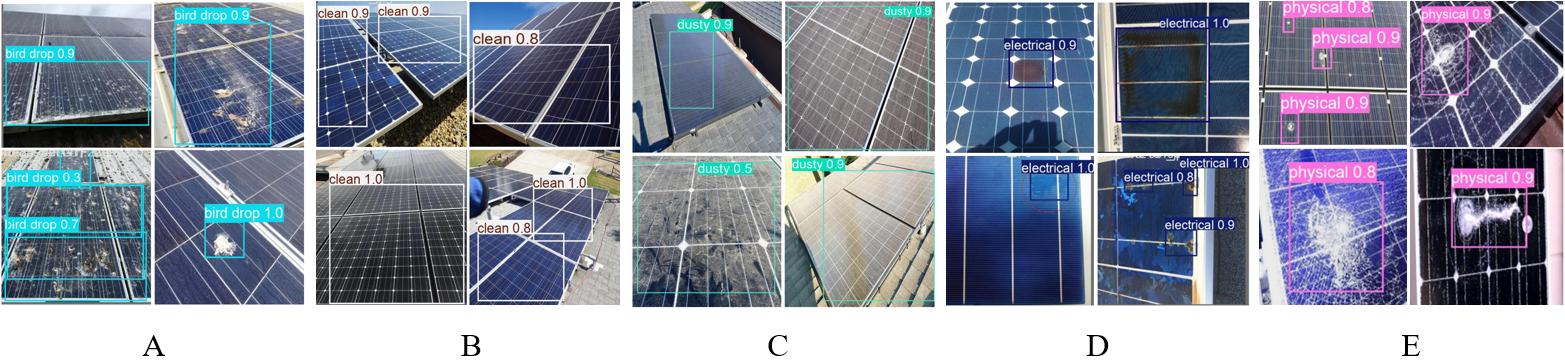}
    \caption{Detection results on solar PV modules with predicted bounding boxes and confidence scores: (A) Bird droppings, (B) Clean panels, (C) Dust accumulation, (D) Electrical defects, and (E) Physical defects.} 
\label{pv-defect-examples} 
\end{figure*}

\begin{figure*}[ht!]
    \centering
    \includegraphics[width=0.9\textwidth]{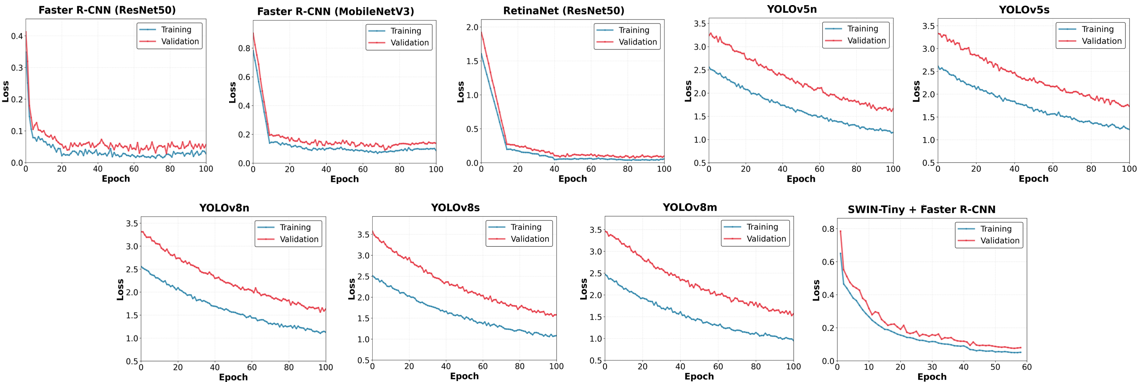}
    \caption{Training and validation loss curves for all evaluated models decrease steadily with minimal divergence, indicating stable learning and generalization to unseen test data.}
    \label{convergence}
\end{figure*}

We present representative qualitative predictions produced by Faster R-CNN with a ResNet50 backbone on the unseen test set, as shown in Fig.~\ref{pv-defect-examples}. This figure illustrates predicted bounding boxes and confidence scores across the five inspection classes, providing a visual assessment of localization behavior and confidence assignment under different realistic imaging conditions. These examples complement the quantitative evaluation by demonstrating how the detector responds to diverse surface appearances and defect indications in real-world scenarios. A large proportion of detections attain IoU values approaching 1, with multiple instances achieving an IoU of exactly 1 (e.g., Bird droppings, Clean panels, Electrical defects), reflecting highly precise and perfect localization. In addition, training and validation loss convergence trends for the evaluated architectures are summarized in Fig.~\ref{convergence}. The loss trajectories decrease steadily, and validation curves closely follow the corresponding training curves, indicating stable optimization under the adopted training configurations.

\section{Conclusions and Future Work}
\label{sec5}

This benchmark demonstrates that architectural design and evaluation priorities strongly influence detector performance for UAV-based solar panel inspection. Two-stage CNN-based detectors consistently deliver superior localization accuracy, while lightweight backbones such as MobileNetV3 enable favorable reliability-efficiency trade-offs for practical deployment. Single-stage detectors exhibit a precision-oriented behavior, making them suitable for applications in which false positives must be minimized. These findings provide clear guidance for selecting detection architectures in real-world photovoltaic inspection systems and establish a reproducible baseline for future research in UAV-based PV defect detection. Future work will focus on expanding the defect dataset with additional field-acquired samples that capture broader environmental and operational variability. Additionally, augmentation strategies tailored specifically for transformer-based architectures could be explored to better leverage their self-attention mechanisms and potentially improve their competitive performance relative to CNN-based detectors.
\bibliographystyle{IEEEtran}
\bibliography{manuscript}











\end{document}